\documentclass[conference,draftcls,onecolumn]{IEEEtran}

\usepackage{amsmath,amssymb,amsfonts}
\usepackage{graphicx}
\usepackage{hyperref}

\def\BibTeX{{\rm B\kern-.05em{\sc i\kern-.025em b}\kern-.08em
    T\kern-.1667em\lower.7ex\hbox{E}\kern-.125emX}}
    
\def\bfx{{\bf x}}
\def\bfy{{\bf y}}
\def\bfz{{\bf z}}
\def\bfX{{\bf X}}
\def\bfY{{\bf Y}}
\def\realR{{\mathbb{R}}}

\makeatletter
\def\blfootnote{\xdef\@thefnmark{}\@footnotetext}
\makeatother

\begin{document}
\bstctlcite{MyBSTcontrol}

\title{Detection Is Tracking: Point Cloud Multi-Sweep Deep Learning Models Revisited}

\author{
\IEEEauthorblockN{Lingji Chen}
\IEEEauthorblockA{chen-lingji@ieee.org}
}

\maketitle
\pagestyle{plain}
\thispagestyle{plain}

\blfootnote{\bf This idea was conceived while the author was working at Motional.}

\begin{abstract}
Conventional tracking paradigm takes in instantaneous measurements such as range and bearing, and produces object tracks across time. In applications such as autonomous driving, lidar measurements in the form of point clouds are usually passed through a ``virtual sensor'' realized by a deep learning model, to produce ``measurements'' such as bounding boxes, which are in turn ingested by a tracking module to produce object tracks. Very often multiple lidar sweeps are accumulated in a buffer to merge and become the input to the virtual sensor. We argue in this paper that such an input already contains temporal information, and therefore the virtual sensor output should also contain temporal information, not just instantaneous values for the time corresponding to the end of the buffer. In particular, we present the deep learning model called MULti-Sweep PAired Detector (MULSPAD\footnote{pronounced ``mousepad'' :) }) that produces, for each detected object, a {\em pair} of bounding boxes at both the {\em end} time and the {\em beginning} time of the input buffer. This is achieved with fairly straightforward changes in commonly used lidar detection models, and with only marginal extra processing, but the resulting symmetry is satisfying. Such paired detections make it possible not only to construct rudimentary trackers fairly easily, but also to construct more sophisticated trackers that can exploit the extra information conveyed by the pair and be robust to choices of motion models and object birth/death models. We have conducted preliminary training and experimentation using Waymo Open Dataset, which shows the efficacy of our proposed method.
\end{abstract}


\IEEEpeerreviewmaketitle

\section{Introduction}
Conventional object tracking has decades of history, in which radar plays a key role. Because of the ``standoff distance,'' an object typically occupies one resolution cell in the radar's field of view, and radar signals, after being preprocessed, Fourier transformed, and threholded, provide range/bearing/range-rate measurements that are considered to correspond to a particular instant of time. The mathematical framework developed to characterize object motion, sensor measurement, and object tracking, can be nominally described as
\begin{equation}
 \begin{split}
 \bfx(t) &= f(\bfx(t-1), \nu(t-1)), \\
 \bfz(t) &= h(\bfx(t), \omega(t)), \\
 \hat{\bfx}(t) &= \phi(\hat{\bfx}(t-1), \bfz(t)),
  \end{split} \label{eqn:radar}
 \end{equation}
 where measurement $\bfz$ is instantaneous without memory, and memory is only in the internal state $\hat{\bfx}$ of the tracker that tries to estimate the object state $\bfx$.

In applications such as autonomous driving, lidar currently plays a key role. With high data rate and short distance to the objects, lidar points reflected from a few hundred objects in the scene easily number in the hundreds of thousands,  and therefore how to make good use of them becomes very challenging. This challenge has been met by neural network based models that turn lidar point clouds into measurements that are easily related to the objects, such as their bounding boxes. We can think of such models as ``virtual sensors.'' If a single sweep of lidar data is used, we can still define ``sensor measurements'' that are instantaneous values, to fit the framework described by Equation~(\ref{eqn:radar}). However, more often than not, in order to enhance signal to noise ratio and improve detection performance, multi-sweep lidar data is used as input to the virtual sensors \cite{caesar2020nuscenes}. Such an input has memory and is no longer instantaneous, and therefore the virtual sensor measurements should have memory too. The measurement equation in (\ref{eqn:radar}) should be replaced by something to reflect this fact, and perhaps a new mathematical framework should be developed accordingly to support the buffered input approach.

In more concrete terms, here is the problem: We put multi-sweep lidar data into a buffer, merge them through a coordinate transformation, and produce bounding boxes corresponding to the time {\em at the end of the buffer}. Purely from the point of view of symmetry, should we not also produce bounding boxes corresponding to the time {\em at the beginning of the buffer}\footnote{We can ask for any time in the buffer, as long as we have the ground truth to learn it.}?

\begin{figure*}[!htb]
 \centering
 \includegraphics[width=0.4\textwidth]
 {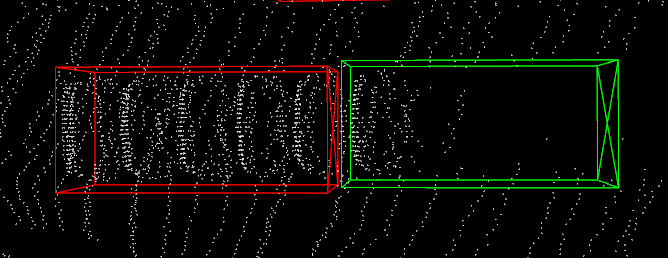}
 \includegraphics[width=0.25\textwidth]
 {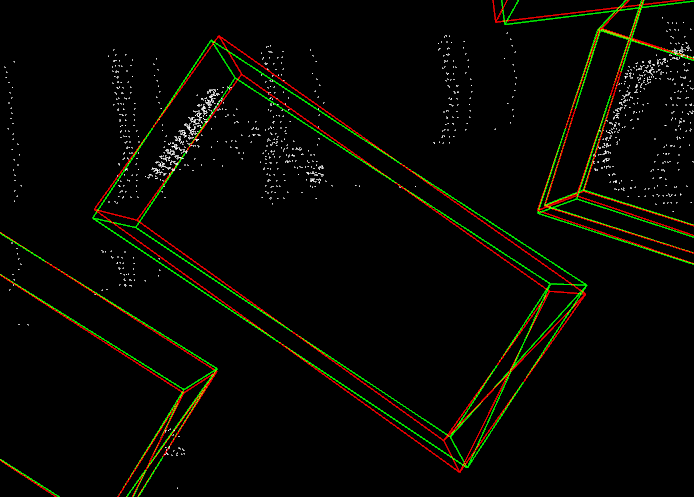}
 \includegraphics[width=0.2\textwidth]
 {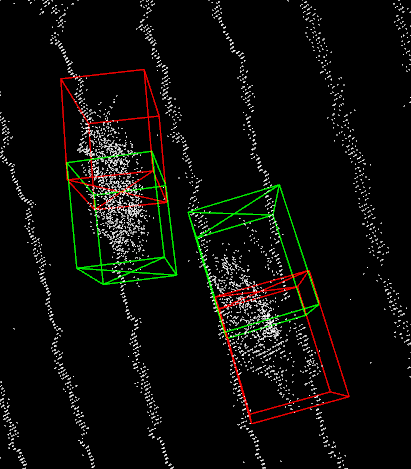}
 \caption{With 6 lidar sweeps, detections are obtained in pairs: a green bounding box at the current sweep indexed by 0, and a red bounding box at the past sweep indexed by -5. Left: a moving car; middle: a parked car; right: two pedestrians walking in opposite directions. } \label{fig:pairs}
\end{figure*}

To obtain symmetry, we have developed the proof-of-concept model MULSPAD, which is based on VoxelNeXT \cite{chen2023voxelnext},  to produce paired detections for each object, as shown in Figure~\ref{fig:pairs} using Waymo Open Dataset \cite{sun2020scalability}. We use 6 sweeps indexed by $[-5, -4, -3, -2, -1, 0]$ where $0$ corresponds to ``current time'' and $-5$ corresponds to ``5 sweeps ago.''

When ground truth provides velocity, using multi-sweep lidar data can learn velocity too, in addition to position information. In our view, learning velocity is subsumed by learning a paired representation, which we believe should make the subsequent tracking task easier and/or more robust.

This paper is organized as follows. In Section~\ref{sec:related} we present a mathematical description of the three algorithms used respectively in CenterPoint \cite{yin2021center}, VoxelNeXT \cite{chen2023voxelnext}, and MULSPAD of this paper, so that it is clear how they are related and how they differ. In Section~\ref{sec:detection} we provide some details of our detector model, and in Section~\ref{sec:tracking} we present the likelihood models used for tracking. Evaluation and ablation studies are presented in Section~\ref{sec:res}, and conclusions are drawn in Section~\ref{sec:con}.

\section{Related Work} \label{sec:related}
Our detector architecture is motivated by VoxelNeXT \cite{chen2023voxelnext} which in turn follows the approach of CenterPoint \cite{yin2021center}, so here we present a somewhat abstracted description in order to better highlight the commonalities and differences.

Let ${\bf X}^0$ be the set of lidar points to be taken in by a deep learning model. If multiple sweeps have been accumulated, then suitable coordinate transformation is carried out so that all points are expressed in the same coordinate frame. The dimension $d_0$ of the vector representing each point is at least three for position $(x, y, z)$; additional dimensions can be for (relative) time, intensity, or even ``painted'' features from other modalities \cite{vora2020pointpainting}. The computation in the model usually takes several stages, with each stage involving a 3D or 2D grid which we will denote by $G^i$ for the $i$-th grid. With an abuse of notation we denote the non-grid, irregular subset by $G^0$ on which ${\bf X}^0$ is defined:
\[
 \bfX^0 = \left\{ \bfx^0_{p^0_i} \in \realR^{d_0}, \; p^0_i \in G^0 \subset \realR^3, i=1, \ldots, N^0 \right\}.
\]

By a process of voxelization, we quantize $G^0$ into $G^1$, and apply a set function $q(\cdot)$ to all the points (after a random shuffling and up to a given limit on the number) falling into the same voxel, to produce a single vector of dimension $d_1$ (which equals $d_0$ in \cite{yin2021center} and \cite{chen2023voxelnext}, and $6 \, d_0$ in our paper for 6 sweeps), so that we have
\[
\bfX^1 = \left\{ \bfx^1_{p^1_i} \in \realR^{d_1}, \; p^1_i \in G^1, i=1, \ldots, N^1 \right\}.
\]

By applying convolution, an output can be produced on a grid whose grid size is a multiple of that of the input, so that we have, for integer $t > 1$,
\[
\bfX^t = \left\{ \bfx^t_{p^t_i} \in \realR^{d_t}, \; p^t_i \in G^t \subset G^{t-1}, i=1, \ldots, N^t \right\},
\]
where in our paper, $d_2 = 32$, $d_3 = 64$, and $d_4 = d_5 = d_6 = d_7 = 128$. The values are of similar magnitude in \cite{yin2021center} and \cite{chen2023voxelnext}.

The notation $G_t \subset G_{t-1}$ has two implications: One is that $G_t$ is a grid on which subsequent computations are defined. The other is that it can be interpreted as some special points in $G_{t-1}$ so that the vectors defined on these points can be collected together with those defined on $G_{t-1}$. We use the superscript notation $m \to n$ to denote such ``reinterpretation'':
\[
 \bfX^{m \to n} \triangleq \left\{ \bfx^m_{p^{m \to n}_i}, \; p^{m \to n}_i \in G^{m \to n}, i=1, \ldots, N^m \right\}.
\]
With these notations, we can say that in VoxelNeXT \cite{chen2023voxelnext} the following set of feature vectors is collected:
\begin{equation} \label{eqn:their_X}
 \bfX = \bfX^4 \cup \bfX^{5 \to 4} \cup \bfX^{6 \to 4}.
\end{equation}

In this paper we add one more stage:
\begin{equation} \label{eqn:X}
  \bfX = \bfX^4 \cup \bfX^{5 \to 4} \cup \bfX^{6 \to 4} \cup \bfX^{ 7 \to 4}.
\end{equation}

Sparsity of the voxel features from lidar point clouds is exploited in the following two ways:
\begin{enumerate}
 \item Regular convolution applied to sparse vectors, implemented in an efficient way. There is ``dilation'' in the sense that a grid point site can have non-zero  ({\em active}) output even if the input is zero ({\em inactive}), due to its active neighbors.
 \item Submanifold sparse convolution \cite{graham20183d} that maintains sparsity. There is no dilation; a site has active output only if it has active input.
\end{enumerate}
The CenterPoint approach \cite{yin2021center} kind of ``abandons'' sparsity mid-way: It first restores the dense format of $\bfX^4$, and then from it produces a set $\bar{\bfX}^4$ through a series of convolutions (performed in the {\em neck} of the model), and another set $\bar{\bar{\bfX}}^4$ through downsampling and upsampling (or, convolution and deconvolution), with cardinality $|\bar{\bfX}^4| = |\bar{\bar{\bfX}}^4| = |G^4|$. Then it concatenates the two to have the feature set
\[
 \bfX = \left[ \bar{\bfX}^4, \bar{\bar{\bfX}}^4 \right].
\]
So far the grid is in 3D. To obtain a bird's eye view, we collect the points with the same $(x, y)$ index in 2D position, and either apply a set function, as is done in VoxelNeXT \cite{chen2023voxelnext} and in this paper, or apply convolution with a stride of 3 in the $z$ index followed by concatenation, as is done in CenterPoint \cite{yin2021center}. To avoid excessive notations we will still call the feature set after the $\mbox{BEV}(\cdot)$ operation, $\bfX$:
\[
 \bfX \triangleq \mbox{BEV} (\bfX).
\]
Each feature vector $\bfx \in \bfX$ is a complicated nonlinear function $f_{\bfx} (\cdot)$ of a set of vectors in $\bfX^0$. If we were to construct a computation graph with source nodes in $\bfX^0$ and sink nodes in $\bfX$ according to the deep learning model just defined, then for each sink node there is a set of source nodes that have paths to reach it. We denote this set by $\mathcal{N}^{-1}(\bfx)$, the ``inverse neighborhood'' of $\bfx$. This is essentially the Effective Receptive Field (ERF) for $\bfx$ \cite{chen2023voxelnext}.

The ground truth for training and evaluation provides a set of objects with their bounding boxes. To construct a deep learning model based on BEV, we choose to ``represent'' an intended detection target by a point in 2D, and maps it to a feature vector to be used as an ``anchor'' or a ``query'' for subsequent classification and regression. More specifically, given a representative $y \in \realR^2$, we map it to a feature $\bfx = \pi(y) \in \bfX$. Let
\[
 \bfY = \left\{ \bfy_k, k=1, \ldots, K \right\} \subset \realR^2
\]
be the set of representatives of $K$ objects given in the ground truth. Then we can define the positive examples
\[
\bfX^+ \triangleq \pi(Y) = \left\{\pi(\bfy_k), \bfy_k \in \bfY, k=1, \ldots, K \right\},
\]
and the negative examples $\bfX^- \triangleq \bfX \backslash \bfX^+$. Thus learning involves two joint problems:
\begin{enumerate}
 \item a classification problem on $\bfX$, through the use of a {\em heatmap} $H(\bfX)$, and
 \item a regression problem on $\bfX^+$, through the use of many {\em heads} of nonlinear regression functions.
\end{enumerate}

To be successful, the choice of anchors should enjoy the following properties (after training has converged):
\begin{itemize}
 \item Each feature vector in $\bfX^+$ is distinct enough for classification.
 \item Each feature vector in $\bfX^+$ contains enough information about the object it represents. Let $\mathcal{N}(\bfy) \subset \bfX^0$ be the set of lidar points that belong to the object represented by $\bfy$. Then typically we want the effective receptive field to cover the entire object, i.e., $\mathcal{N}(\bfy) \subset \mathcal{N}^{-1}(\pi(\bfy))$.
\end{itemize}
In CenterPoint \cite{yin2021center}, $\bfy$ is chosen as the bounding box center, and because the heatmap $H(\bfX)$ is dense, $\pi(\bfy)$ picks the corresponding quantized grid point. In VoxelNeXT \cite{chen2023voxelnext} $\bfy$ is still chosen as the bounding box center, but the heatmap is sparse (because $\bfX$ in Equation~(\ref{eqn:their_X}) is), so $\pi(\bfy)$ picks the {\em closest} grid point with a heatmap feature vector on it. Since most lidar points fall on the exterior of an object, the closest to the center may still be some distance away, which was lucidly illustrated in \cite{chen2023voxelnext}. In this paper MULSPAD produces paired detections, so $\bfy$ is chosen as the {\em midpoint} between the two bounding box centers, and $\pi(\bfy)$ picks the {\em closest} grid point with a heatmap feature vector on it. Given the multi-sweep nature, the midpoint is not necessarily in the interior of an object, so it may have some lidar points nearby. We mention in passing that another choice for $\bfy$ (especially for single-sweep approaches) is the bounding box corner that is closest to the coordinate frame origin where the lidar sensor sits; this usually corresponds to the corner with the densest lidar points.

\section{Detection} \label{sec:detection}
In this section we provide some details of MULSPAD that have not been discussed so far.

\subsection{Architecture}
\begin{figure*}[!htb]
 \centering
 \includegraphics[width=0.98\textwidth]
 {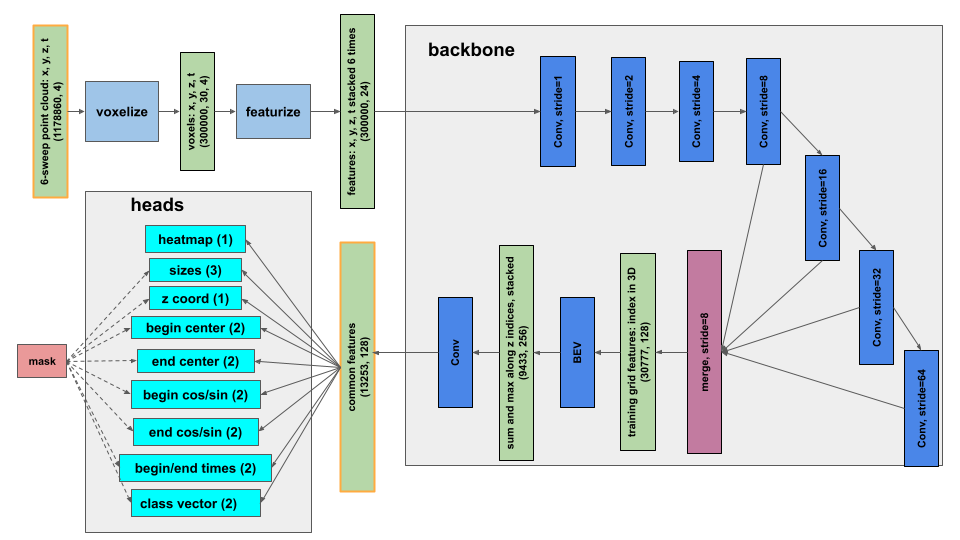}
 \caption{Architecture of MULSPAD, motivated by \cite{chen2023voxelnext}. } \label{fig:arch}
\end{figure*}

Our model is an extension of VoxelNeXT \cite{chen2023voxelnext} that takes full advantage of the sparsity of the lidar points. The architecture of MULSPAD is shown in Figure~\ref{fig:arch} where, {\em for a particular batch}, dimensions of feature vectors at various stages of processing are marked.

Since we are using multiple sweeps, we increase the budget for the number of voxels used. For each voxel, since we are using 6 sweeps and therefore have 6 distinct relative time values that can be attached to a lidar vector, we allow up to 30 points in a voxel, and take the mean for each time value. We then stack up the 6 mean vectors to form a feature vector for the voxel.

Paired detections occupy a bigger span of space, so to increase the effective receptive field $\mathcal{N}^{-1}(\cdot)$, we add another stage $\bfX^7$ in (\ref{eqn:X}). In forming BEV, for vectors falling on the same $(x, y)$ index, we stack up their sum and their elementwise max to form the feature vector; in VoxelNeXT \cite{chen2023voxelnext} only the mean is used.

We use a single, mixed-class heatmap instead of three heatmaps (one per class), but add a head to regress object types as shown in Figure~\ref{fig:arch}. For faster training and proof-of-concept experimentation we also ``thin out'' the {\tt Conv} blocks in Figure~\ref{fig:arch} to have fewer internal layers. In regression we let the pair of bounding boxes share the same length, width, and height, and the same $z$ coordinate value, but have their own center locations and headings. The begin and end times of the bounding boxes are a little involved and will be presented next.

\subsection{Begin and end times}
\begin{figure*}[!htb]
 \centering
 \includegraphics[width=0.2\textwidth]
 {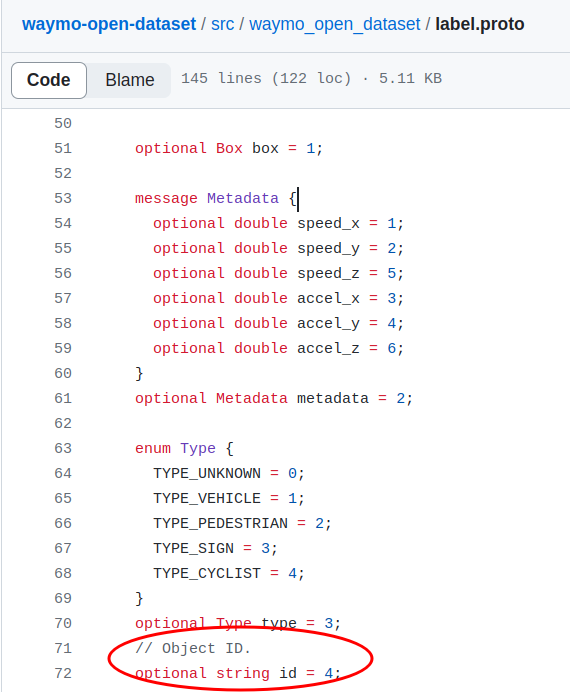}
 \caption{ In Waymo Open Dataset, ground truth track IDs such as {\tt 3uxz-RltuMwlmgDehBaZfA} are stored in the field {\tt id} inside {\tt Label}.} \label{fig:label}
 \end{figure*}

First we note that to construct MULSPAD we need not only ground truths as detections, but also ground truths as tracks. Waymo Open Dataset provides ground truth track IDs in {\tt Label}; See Figure~\ref{fig:label}.

\begin{figure*}[!htb]
 \centering
 \includegraphics[width=0.5\textwidth,trim=70 10 90 20,clip]
 {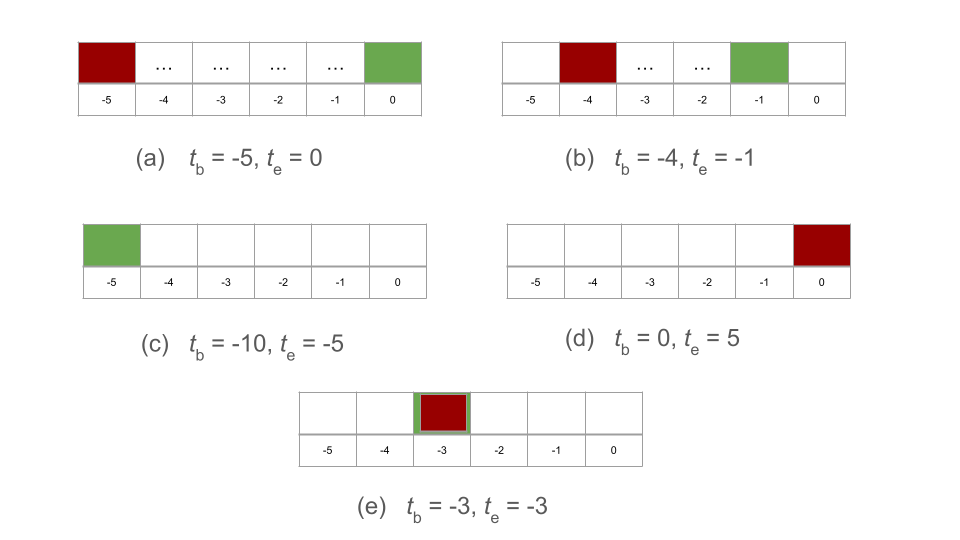}
 \caption{ Define the begin time $t_b$ and the end time $t_e$ of a ground truth track in the 6-sweep buffer. Most cases fall into (a) while some into variations of (b); the complication of ``breakage'' is ignored in this paper. To distinguish from the ``singleton'' case in (e), a ``birth time target'' is used for (c), and a ``death time target'' is used for (d). } \label{fig:def_times}
\end{figure*}

For each ground truth track, we examine its bounding boxes inside the sweeps under consideration. Most cases are like (a) in Figure~\ref{fig:def_times}, where all 6 bounding boxes are present. In this paper we only look at the begin and end times and pretend that it is contiguous in between. In other words, we ignore the complication of breakage and leave it for future work. There are thus  cases where
\begin{itemize}
 \item the start happens ``in the middle,'' which indicates object birth, or object coming out of occlusion, or
 \item  the end happens ``in the middle,'' which indicates object death, or object going into occlusion.
\end{itemize}
Because of such implications, the case (e) shown in Figure~\ref{fig:def_times} where the ``singleton'' happens in the middle should be distinguished from the singleton cases in (c) and (d). When we cannot ``see'' the birth, we use an artificial ``birth time marker'' as the target of learning. Likewise for death.

\section{Tracking} \label{sec:tracking}
In this section we discuss tracking using the paired detections from MULSPAD as input. When temporal information is available and it tells us not only where each object {\em is}, but also where it {\em was}, then conceptually tracking is already done. In fact, with some simple heuristics, a rudimentary tracker can be constructed that performs reasonably well.

\begin{figure*}[!htb]
 \centering
 \includegraphics[width=0.5\textwidth,trim=100 25 100 25,clip]
 {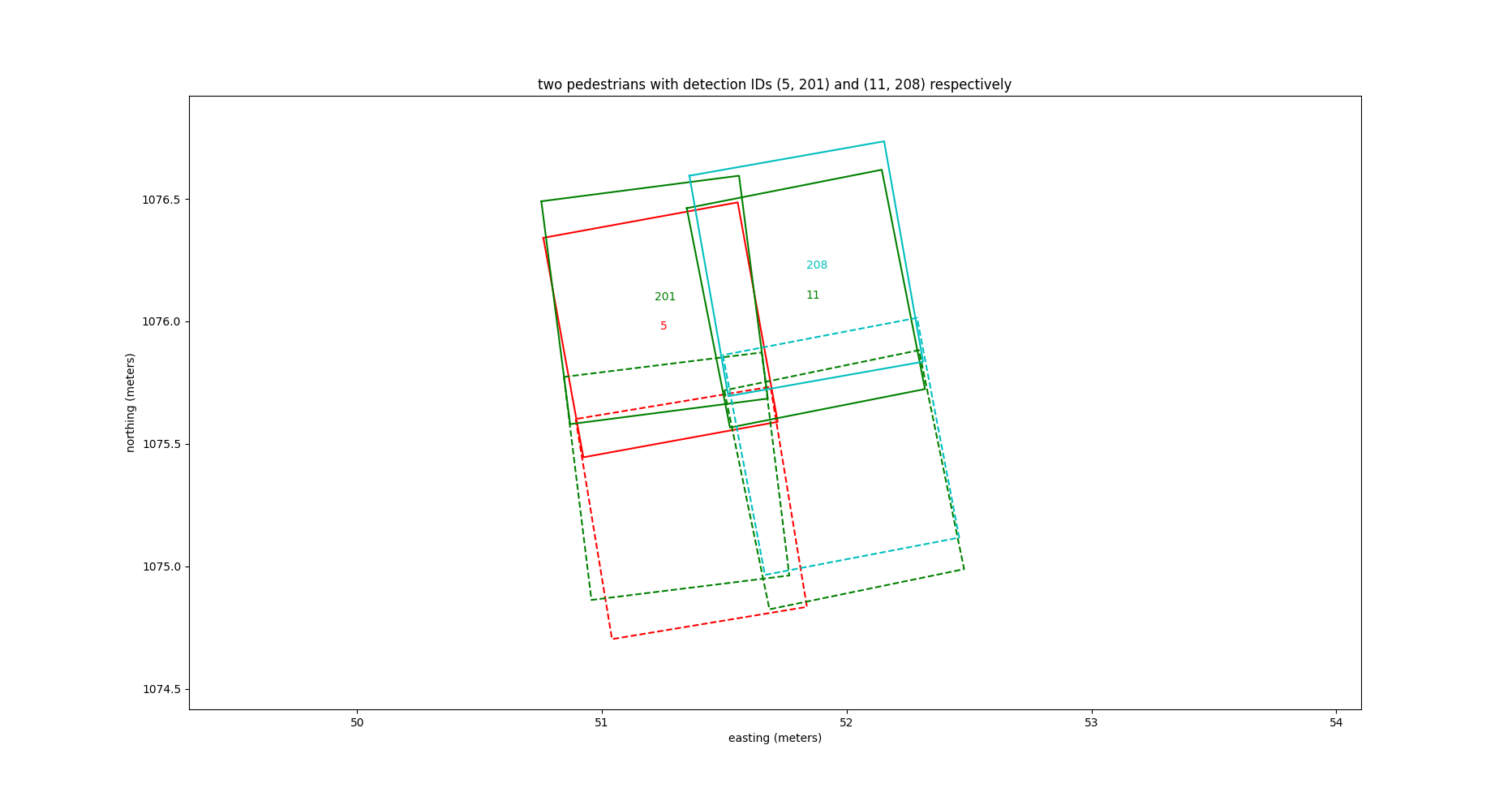}
 \caption{ Two pedestrian tracks with their constituent detections. The pair has the same color, and Detection ID is printed in the center of the solid bounding box. } \label{fig:two_ped_trks}
\end{figure*}

We argue that the richer information provided by the paired detections can make the tracker more robust, and/or can be taken in by a more sophisticated tracker for enhanced performance. Figure~\ref{fig:two_ped_trks} shows two pedestrian tracks, where detection pairs are shown with solid and dotted boxes in the same color. We can infer that this tracking result would not be sensitive to estimation inaccuracies in velocity or heading because we are ``seeing'' directly where the pedestrians are coming from.

\subsection{Random Finite Set (RFS) tracker, special case}
\begin{figure*}[!htb]
 \centering
 \includegraphics[width=0.7\textwidth,trim=10 80 50 0,clip]
 {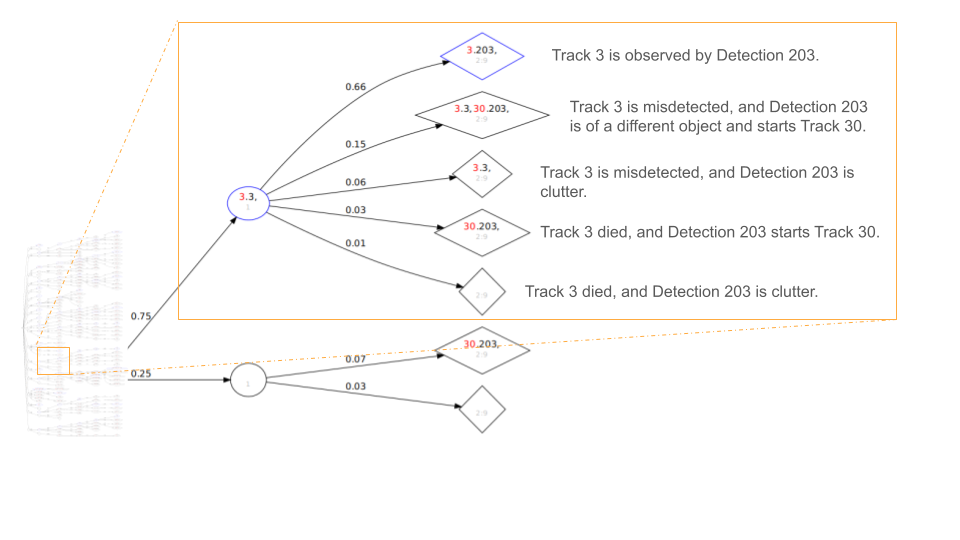}
 \caption{Hypothesis pedigree of an RFS tracker (recorded in debug mode), with a zoom-in that gives explanation to the hypotheses. For figure and algorithm details, see \cite{chen2022merge}.} \label{fig:trk3}
\end{figure*}

\begin{figure*}[!htb]
 \centering
 \includegraphics[width=0.5\textwidth]
 {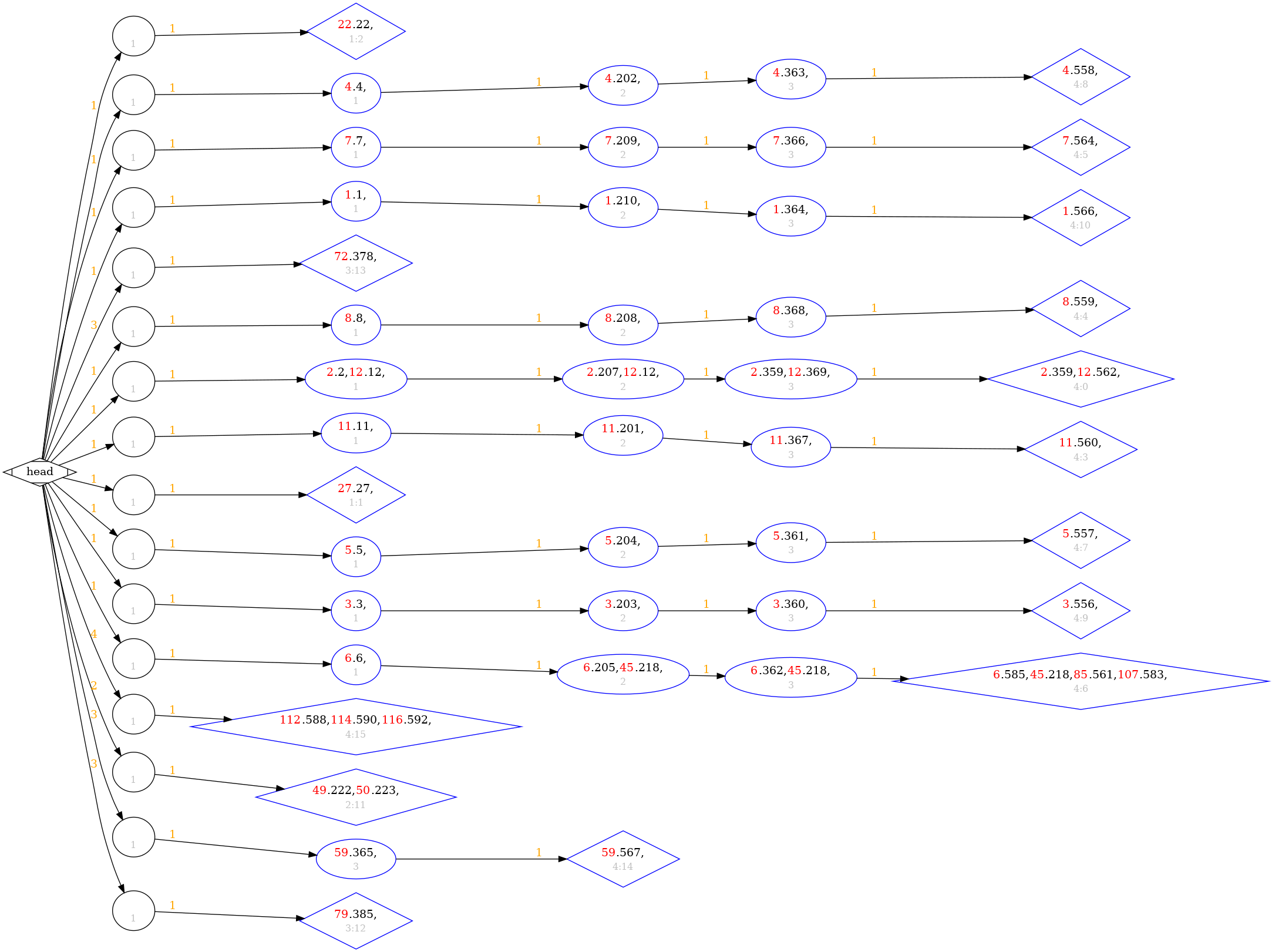}
 \caption{Hypothesis pedigree of an RFS tracker (recorded in debug mode), with the hypothesis budget set to 1 to get a greedy assignment tracker as a baseline. For figure and algorithm details, see \cite{chen2022merge}.} \label{fig:hypo1}
\end{figure*}

We want to establish a baseline tracker (along the line of \cite{weng20203d}) that does essentially greedy assignment even though it  ingests paired detections produced by MULSPAD. This is achieved by first modifying an existing RFS tracker that performs multiple hypothesis reasoning, and then setting the budget parameter for hypotheses to be exactly $1$, so as to get the special case of a tracker with greedy assignment, i.e., assigning measurements to tracks and committing to this assignment for subsequent processing. To provide a wider context, in Figure~\ref{fig:trk3} we first illustrate the general case and describe the new hypotheses generated when a new detection arrives. In Figure~\ref{fig:hypo1} we show the special case baseline tracker that is mathematically equivalent to greedy assignment. However, the Merge/Split algorithm \cite{chen2022merge} recently developed for the RFS framework \cite{vo2014labeled} makes the implementation more efficient and versatile.

We mention in passing that, even though an independence assumption between tracks is made in order for an RFS tracker to efficiently enumerate hypotheses, the tracker can in certain cases \cite{chen2021multi} consider tracks jointly without losing that efficiency, e.g., the RFS tracker can efficiently rule out impossible hypotheses that contain collision or spatial incompatibility of tracks.

\subsection{Likelihood models}
Defining some negative log likelihood ratio to be cost, the RFS framework employs a cost matrix and the Murty's algorithm \cite{murty1968algorithm} to enumerate hypotheses in descending order of their likelihoods/weights; see Figure~\ref{fig:trk3} where weights are shown on the arcs. As proof-of-concept, in this paper we adopt some rudimentary likelihood models to construct our tracker that ingests paired detections.
\subsubsection{Static and slow moving objects}
We use a circular buffer to store past detections in a track. For static and slow moving objects, we identify a bounding box in the track and a bounding box in the new detection that are closest in time but not necessarily at the same time, and they may already overlap. The point is that in this model we are {\em not} using motion prediction. Let $\lambda$ be their 2D Intersection Over Union (IOU). Define a threshold parameter $\lambda_0$. We assume a likelihood model
\[
 \mbox{if } \lambda \ge \lambda_0: 1 - \lambda \sim N(0, \sigma_\lambda^2),
\]
where $N(\mu, S)$ denotes a normal distribution with mean $\mu$ and covariance $S$, and $\sigma_\lambda$ is a design parameter.

\subsubsection{Moving objects}
\begin{figure*}[!htb]
 \centering
 \includegraphics[width=0.8\textwidth,trim=100 200 110 80,clip]
 {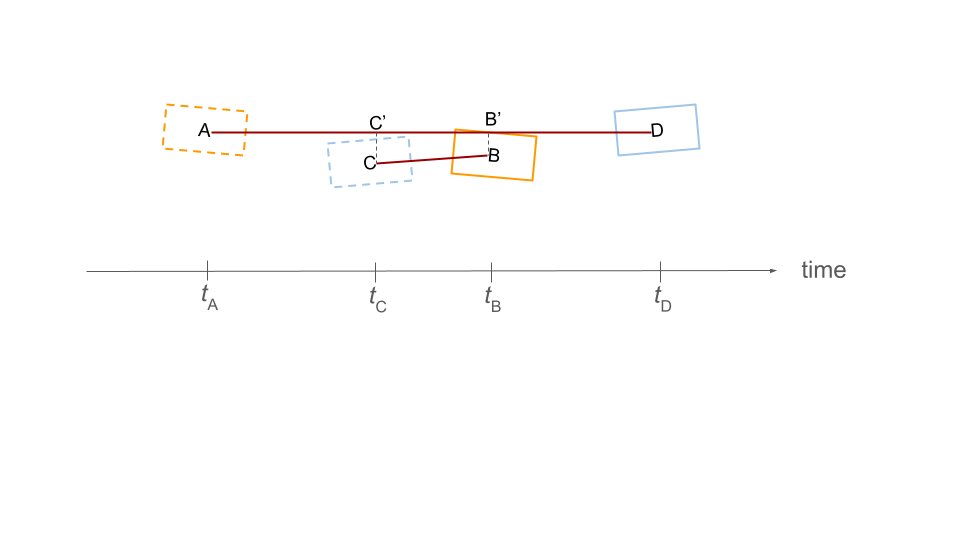}
 \caption{ Building simpler likelihood models based on intersection $|C'B'|$ over union $|AD|$, and on maximum lateral deviation $|CC'|$ . } \label{fig:iou}
\end{figure*}

If $\lambda < \lambda_0$, then we adopt the following likelihood model, which can be considered a heuristic, 1D, ``poor man's IOU.'' We pick four bounding boxes, two from the track, and two from the new paired detection, as illustrated in Figure~\ref{fig:iou}.

Define
\[
 \lambda = \frac{|C'B'|}{|AD|}, \; \lambda_t = \frac{|t_B - t_C|}{|t_D - t_A|},
\]
where $| \cdot |$  denotes line segment length for $\lambda$ and absolute value for $\lambda_t$ respectively, and we assume a likelihood model
\[
 \lambda - \lambda_t \sim N(0, \sigma_t^2),
\]
where $\sigma_t$ is a design parameter. We jointly use another likelihood model for the lateral deviation:
\[
 \max(|BB'|, |CC'|) \sim N(0, \sigma_l^2),
\]
where $\sigma_l$ is a design parameter. There are also design parameters that make the likelihoods of all hypotheses ``in the same unit'' or ``dimensionless'' \cite{bar2007dimensionless}.

\subsection{Some tracking results}
\begin{figure*}[!htb]
 \centering
 \includegraphics[width=0.95\textwidth]
 {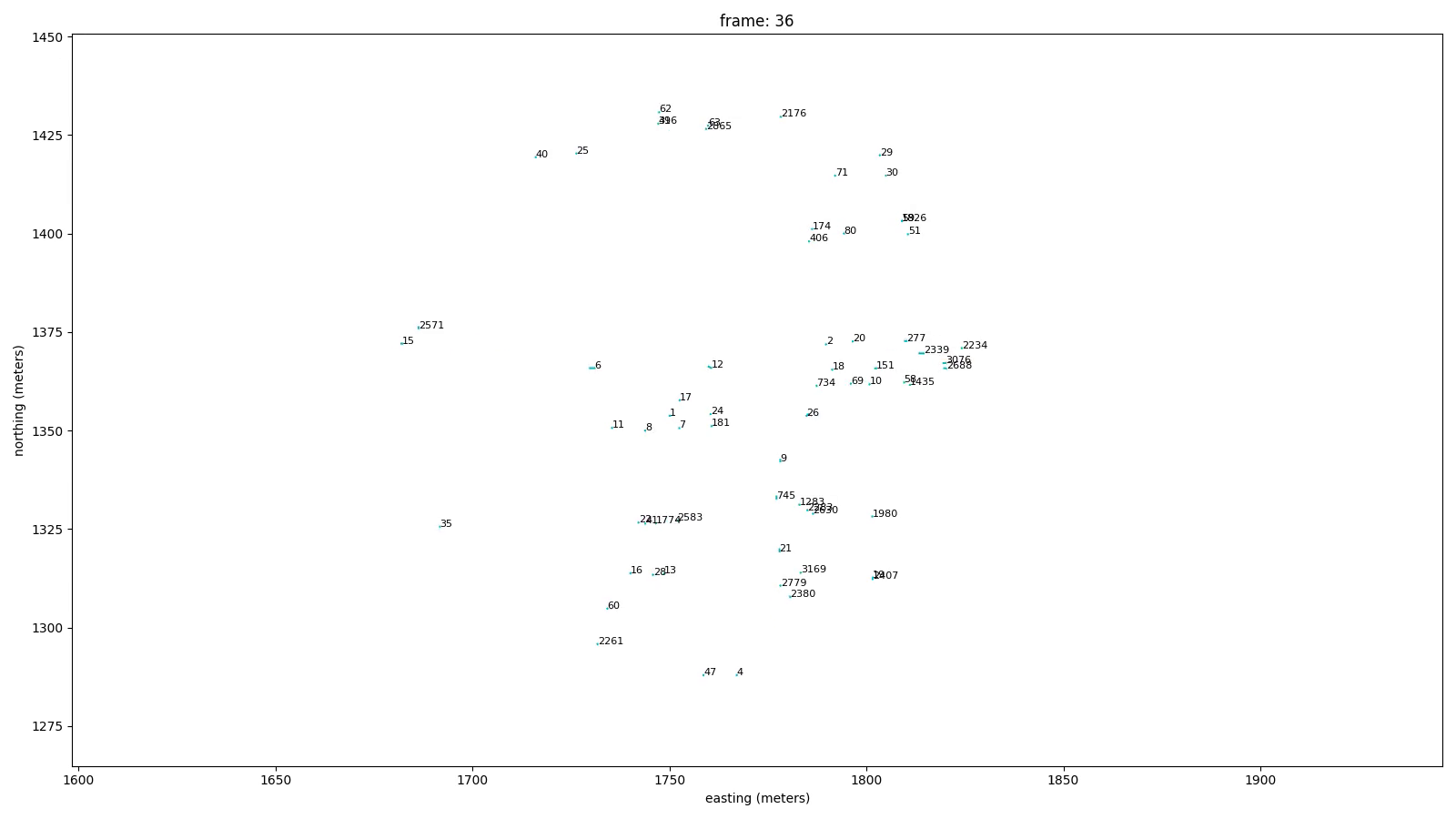}
 \caption{A snapshot of tracking result: Track IDs are printed at object centers with attached velocity vectors at 1/10th their magnitude (for data rate at 10 Hz). For an animation of all frames in the 20-second segment, see \url{https://github.com/lambertch/MULSPAD/blob/main/README.md}} \label{fig:snapshot}
\end{figure*}

We use the validation set in Waymo Open Dataset 3D Tracking Challenge to gauge our tracking performance, with help from published repositories such as \cite{pang2022simpletrack} for data format conversions etc. A snapshot of tracker results is shown in Figure~\ref{fig:snapshot}. The efficacy of our proposed approach is confirmed.

\section{Results and Ablation Studies} \label{sec:res}
With preliminary training and experimentation and the baseline RFS tracker, we achieved a MOTA/L2 of $0.577$ for near range vehicles on the validation set in Waymo Open Dataset 3D Tracking Challenge.

Work is still on going, and results and ablation studies will be reported in an update of this paper.

\section{Conclusions} \label{sec:con}
In this paper we explore a new perspective on the temporal nature of the input when multi-sweep lidar data is fed into a neural network detection model. We have presented MULSPAD that produces for each detected object a pair of bounding boxes, one for the current time, and one for a past time. This can be considered a ``symmetry completion'' for the typical approach that produces bounding boxes only for one end of the input buffer but not the other end. It involves straightforward changes to a detection model and requires only marginal additional computation, so there is no reason not to do it in a multi-sweep approach. We hope that this perspective will open up new possibilities for detection and tracking.

\section*{Acknowledgment}
This idea was conceived while the author was working at Motional. He would like to thank Dr. Sourabh Vora for helpful discussions. He would also like to thank Scott Ettinger and Alex Gorban from the Waymo Open Dataset team for answering questions.

The author would like to thank Ms. Hua Yang for her unique support: This paper would not exist without you.

\bibliographystyle{IEEEtran}
\bibliography{MyBSTcontrol,det_trk}

\end{document}